\documentclass[a4paper, 10pt, conference]{ieeeconf}      

\IEEEoverridecommandlockouts                              

\usepackage[T1]{fontenc}
\usepackage{graphicx}
\usepackage{multicol}
\usepackage{dblfloatfix}
\usepackage{epsfig} 
\usepackage{amsmath} 
{
	
}
\usepackage{amssymb}  
\usepackage{subfig}
\usepackage{float}
\usepackage{hyperref}
\usepackage{cleveref}
\Crefformat{figure}{#2Fig.~#1#3}

\usepackage{color}

\usepackage[sc]{mathpazo}

\usepackage[font=footnotesize,labelfont=bf]{caption}
\usepackage{cite}
\usepackage{mathrsfs}
\usepackage{algorithm, algorithmic}
\usepackage{bm}
\usepackage{makecell}

\renewcommand{\baselinestretch}{1}
\title{\LARGE \bf
	Interpretable Modelling of Driving Behaviors in Interactive Driving Scenarios based on Cumulative Prospect Theory
}

\author{Liting Sun, Wei Zhan, Yeping Hu and Masayoshi Tomizuka
	\thanks{*This work was partially supported by Mines ParisTech Foundation, ``Automated Vehicles - Drive for All''. }
	\thanks{Liting Sun, Wei Zhan, Yeping Hu and Masayoshi Tomizuka are with the Department of Mechanical Engineering, University of California at Berkeley, CA, 94720, USA.
		{\tt\small \{litingsun, wzhan, yeping\_hu, tomizuka\}@berkeley.edu}}%
}

\begin{document}

\maketitle
\thispagestyle{empty}
\pagestyle{empty}

\begin{abstract}
Understanding human driving behavior is important for autonomous vehicles. In this paper, we propose an interpretable human behavior model in interactive driving scenarios based on the cumulative prospect theory (CPT). As a non-expected utility theory, CPT can well explain some systematically biased or ``irrational'' behavior/decisions of human that cannot be explained by the expected utility theory. Hence, the goal of this work is to formulate the human drivers' behavior generation model with CPT so that some ``irrational'' behavior or decisions of human can be better captured and predicted. Towards such goal, we first develop a CPT-driven decision-making model focusing on driving scenarios with two interacting agents. A hierarchical learning algorithm is proposed afterwards to learn the utility function, the value function, and the decision weighting function in the CPT model. A case study for roundabout merging is also provided as verification. With real driving data, the prediction performances of three different models are compared: a predefined model based on time-to-collision (TTC), a learning-based model based on neural networks, and the proposed CPT-based model. The results show that the proposed model outperforms the TTC model and achieves similar performance as the learning-based model with much less training data and better interpretability.
\end{abstract}
\section{INTRODUCTION}
\label{sec:intro}

Modelling interactive behavior of human drivers is extremely important to enable safe and full autonomy for vehicles. It not only can facilitate better prediction of human drivers' intentions and motions, but also can serve as valuable asset to generate more human-like decisions and trajectories for autonomous vehicles. 

In the past decade, a great amount of effort has been devoted to driver behavior modelling, e.g., \cite{abuali_driver_2016, wang_modeling_2014, rahman_review_2013}. Most of the proposed methodologies can be categorized into three groups: 1) predefined models, 2) learning models, and 3) utility-driven models. Predefined models generate driving behavior based on IF-THEN rules \cite{mcdonald_development_1997}, or selected key indices such as time-to-collision (TTC) and time-to-intersection (TTI) \cite{kiefer_status_2005}, or some analytical functions dedicated to describe the behavior in specific scenarios. Examples include the intelligent driver model (IDM) \cite{treiber_congested_2000} for car following and the minimizing overall braking model for lane changing (MOBIL) \cite{kesting_general_2007}. Such predefined models are highly interpretable, i.e., explicit physical meanings can be found for all the model structures, variables and parameters. However, these models typically require lots of manual work in designing structures and tuning parameters, which can be overwhelming tasks when the amount of data is large.

Learning models generate driving behavior based on trained machine-learning models. They can be either discriminative models such as support vector machines (SVM) \cite{aoude_driver_2012} and mixture density network (MDN) \cite{hu_probabilistic_2018}, or generative models such as hidden Markov models (HMM) \cite{li_generic_2018IV}, generative adversarial networks (GAN) \cite{li_coordination_2019, gupta2018social} and variational auto-encoder (VAE) \cite{hu_framework_2018, ma2018trafficpredict, Hu2019IV}. Compared to the predefined models, such learning models can better approximate the complicated distributions of human behavior in massive driving data without manual tuning of model parameters. However, they also suffer from several fundamental problems. First, most of the learning models, particularly the deep networks, are data-hungry. With a relatively small amount of data, they can hardly achieve satisfactory performance. Even with sufficient amount of data supplied, they still suffer from a second problem: the lack of causality and interpretability of the learned behavior model. Consequently, it is hard to efficiently generalize them to new scenarios such as those with a varying number of agents or new driving maps.

Utility-driven models come from the theory of mind (TOM) \cite{premack1978does}. A key feature of these models is that they leverage the fact that human drivers are not random agents, but agents who optimize some utility functions. Hence, they assume that human drivers try to make decisions or plan trajectories that maximize their utilities (or minimize the costs). Such assumption is often known as the Boltzmann noisily rational model \cite{baker2014modeling}. 
Stemming from TOM, utility-driven models provide causality inherently, and are more interpretable since all the utilities and constraints are associated with explicit physical meanings, as mentioned in \cite{driggs-campbell_integrating_2017}. In order to infer the various utility functions of human from actual driving data, inverse optimal control (or inverse reinforcement learning (IRL)) \cite{abbeel2004apprenticeship, ziebart2008maximum, Levine2012ICML} has been well adopted. For instance, \cite{sun_courteous_2018} use IRL to model the courteous behavior. \cite{sun_probabilistic_2018} and \cite{Hu2019ITSC} use it for probabilistic reaction prediction under social interactions. Benefiting from causal and interpretable structure design, the utility-driven models are more data-efficient, i.e., a satisfactory model can be learned with a relatively small amount of data. Hence, they provide a promising balance between interpretability, model flexibility, and data-efficiency.

One remaining challenge for the utility-driven models is that, as mentioned above, most of them assume the rationality (or at least noisy rationality) of human drivers with respect to the expected utility theory (EUT). However, there have been substantial evidences in various domains contradicting such assumption. Human behavior is often found to be systematically deviating from the optimal (or rational) behavior predicted by EUT. Examples can be found as framing effect, risk-seeking behavior, loss-aversion behavior, and so on \cite{kahneman1979prospect, tversky1992advances}. In this paper, we define such systematically biased behavior from EUT as ``\textit{irrational behavior}''. In driving scenarios, such irrational behavior can be well observed, particularly when the drivers are interacting with each other. Under such circumstances, the traditional EUT-based utility-driven models can no longer correctly predict the human behaviors, which might cause collisions for the autonomous vehicles. 

Therefore, in this paper, {we aim to extend the utility-based behavior generation model to capture both the rational and irrational behavior of human drivers}. Towards this goal, we reformuate the utility-based models in the framework of the cumulative prospect theory (CPT) \cite{tversky1992advances} -  a well-known non-expected utility theory (NEUT) that can explain many of the irrational behaviors mentioned above. Afterwards, a hierarchical learning algorithm is proposed to learn the utility function, the value function, and the decision weighting function in the developed CPT model. A case study for roundabout merging is presented with real data from the INTERACTION dataset \cite{Wei2019IROS, zhan_2019}. Prediction performances of three different models are compared: a predefined model based on TTC, a learning-based model based on a neural network, and the proposed CPT-based model. The results show that the proposed model outperforms the TTC model, and achieves similar performance as the learning-based method with much less training data and better interpretability.

\section{A BRIEF INTRODUCTION TO UTILITY THEORIES}
\label{sec:utility_theory}
We briefly review two utility theories for modelling the decision-making process of human: the expected utility theory and the cumulative prospect theory, a non-expected utility theory.
\subsection{Expected utility theory}
\label{subsec:expected_utility_theories}
The expected utility theory (EUT) \cite{risk1954exposition} was first introduced by Bernoulli in 1738. It approximates decision makers as maximizers of their expected utilities. Mathematically, the process can be modelled as follows.

Let $\{\bm{a}\}{=}\{a_1, a_2, {\cdots}, a_n\}$ be a set of $n$ possible actions/choices. With each action $a_i$, define the possible state set as $\bm{\{x_i\}}{=}\{{x_{i,1}}, {\cdots}, {x_{i,m}}\}$ with ${x_{i,j}}{\in}\mathcal{R}^k$ for $i{=}1,{\cdots}, n$ and $j{=}1,{\cdots}, m$. The probability of each state is represented by ${p_{i,j}}{=}p({x_{i,j}})$ satisfying $\sum_j^{m} {p_{i, j}}{=}1$. Define $u({x_{i,j}}, a_i)$ as the function that assigns utility to each pair of state and action. Then, under each decision choice $a_i$, the possible outcome profile (i.e., the prospect) can be represented by $\bm{P_i}{=}\left(\bm{u}(a_i), \bm{p_i}\right)$, where $\bm{u}(a_i){=}[u({x_{i,1}}, a_i), u({x_{i,2}}, a_i), {\cdots}, u({x_{i,m}}, a_i)]^T$ is the utility vector defined on the possible state set, and $\bm{p_i}{=}[p({x_{i,1}}), p({x_{i,2}}), {\cdots}, p({x_{i,m}})]^T$ is the corresponding probability vector of $\{\bm{x_i}\}$.

The expected utility $U$ of each decision can then be written as
\begin{IEEEeqnarray}{rCl}
	U(a_i) = U(\bm{P_i})  = \sum_{j=1}^m u({x_{i,j}}, a_i)p({x_{i,j}}),\label{eq:expected_utility}
\end{IEEEeqnarray}
and decision makers choose the action that generates the maximum expected utility, i.e., 
\begin{equation}
a^*{=}\arg \max_{a_i{\in}\{\bm{a}\}}\{U(a_i)\}{=}\arg \max_{a_i{\in}\{\bm{a}\}}\{U(a_1), {\cdots}, U(a_n)\}.\label{eq:EUT-decision}
\end{equation}

Although the EUT has been adopted in many application domains as the dominant model to describe how individuals make decisions under uncertainties, there have been substantial evidences showing that human behavior often violates the EUT hypothesis in a systematic way such as loss aversion, risk seeking and nonlinear preferences \cite{tversky1992advances, allais1953rational}. 

\subsection{Cumulative prospect theory, a non-expected utility theory}
\label{subsec:non_expected_utility_theories}
Many non-expected utility theories (NEUT) were developed to explain the above-mentioned behaviors which deviate from EUT. Among them, the cumulative prospect theory (CPT), proposed by Kahneman and Tversky \cite{tversky1992advances},  is one study that formulates many such biased or irrational human behaviors in a uniform way. Compared to the EUT in (\ref{eq:expected_utility}), CPT introduced two additional concepts in the definition of prospect $\bm{P}$: a value function $v$ defined on the utility and a decision weight function $\pi$ defined on the cumulative probability. Each action is evaluated by the function
\begin{IEEEeqnarray}{rCl}
	V(a_i)&=&V(\bm{P_i})\nonumber\\
	&=&\sum_{j=1}^m v\left(u^{+}(x_{i,j}{,}a_i)\right)\pi_j^{+}{+}v\left(u^{-}(x_{i,j}{,} a_i)\right){\pi}_j^{-},\label{eq:CPT}
\end{IEEEeqnarray}
where the function $v: \mathcal{R}{\rightarrow}\mathcal{R}$ is a strictly increasing function, and $u^{+}(\cdot)$ and $u^{-}(\cdot)$ represent, respectively, the gains and losses of $u(\cdot)$ compared to a reference utility $u_0$. The decision weights are defined as
\begin{IEEEeqnarray}{rCl}
	\pi^+_m&{=}&w^+\left(p(x_{i,m})\right), \quad \pi^-_m{=}w^-\left(p(x_{i,m})\right),\label{eq:decision_weight_1}\\
	\pi^{+}_j &{=}& w^{+}\left(\sum_{k=j}^{m}p(x_{i,k})\right){-}w^{+}\left(\sum_{k=j+1}^{m}p(x_{i,k})\right){,}\label{eq:decision_weight_2}\\
		\pi^{-}_j &{=}& w^{-}\left(\sum_{k=j}^{m}p(x_{i,k})\right){-}w^{-}\left(\sum_{k{=}j{+}1}^{m}p(x_{i,k})\right){,} \nonumber\\
		&&\forall j{=}1,{\cdots}, m{-}1\label{eq:decision_weight_3}
\end{IEEEeqnarray}
where $w^{\pm}{:} [0,1]{\rightarrow}[0,1]$ are both strictly increasing functions with $w^+(0){=}w^-(0){=}0$, and $w^+(1){=}w^-(1){=}1$. 

\begin{figure}[t!]
	\begin{center}
		\includegraphics[width=0.95\linewidth]{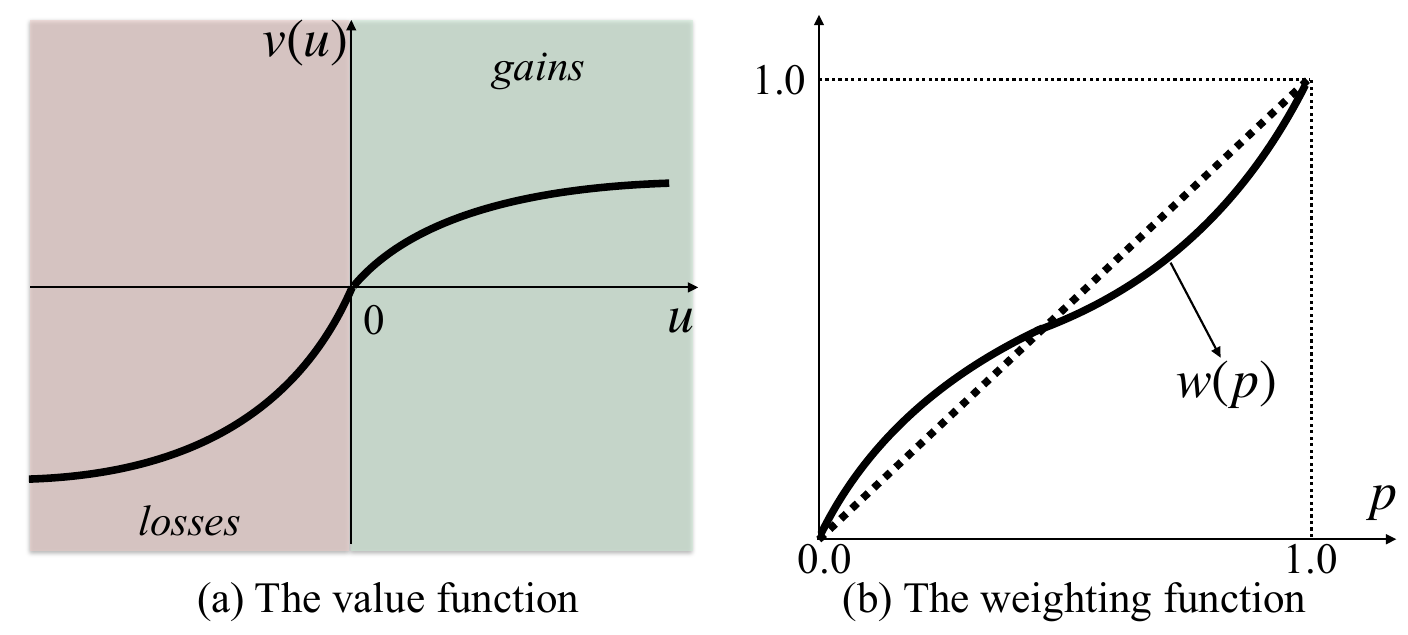}
		\caption{Examples of the value function and weighting function}
		\label{fig: example_value_weight_functions}
	\end{center}
\end{figure}

Typically, the value function $v(u)$ is convex when $u{\ge}u_0$ (gains) and concave when $u{<}u_0$ (losses), and it is steeper for losses than for gains. Figure \ref{fig: example_value_weight_functions}(a) shows one example of the value function when $u_0{=}0$ is set as the reference utility. Many experiment studies have showed that representative functional forms for $v$ and $w$ can be written as
 \begin{IEEEeqnarray}{rCl}
 v(u)&=&\left\{
\begin{aligned}
	& (u-u_0)^\alpha, &\text{if }u\ge u_0 \\
	& -\lambda(u_0-u)^\beta, &\text{if }u<u_0
\end{aligned}
\right.\label{eq:functional_form_v}\\ 	
w^+(p) &=& \frac{p^\gamma}{\left(p^\gamma+(1-p)^\gamma\right)^{1/\gamma}}, \\
w^-(p) &=& \frac{p^\delta}{\left(p^\delta+(1-p)^\delta\right)^{1/\delta}}.\label{eq:functional_form_w}
 \end{IEEEeqnarray}
respectively, with $\alpha, \beta, \gamma, \delta{\in}(0,1]$ and $\lambda{\ge}1$. As shown in \Cref{fig: example_value_weight_functions}(b), such decision weight functions can describe the well-observed behaviors that human tends to over-estimate the occurrence of low-probability events but under-estimate that of the high-probability ones.

Similary to EUT, the CPT model assumes that the decision makers choose the action that yields the maximum value defined in (\ref{eq:CPT}), i.e.,
\begin{equation}
a^*{=}\arg \max_{a_i{\in}\{\bm{a}\}}\{V(a_i)\}{=}\arg \max_{a_i{\in}\{\bm{a}\}}\{V(a_1), {\cdots}, V(a_n)\}.\label{eq:CPT-decision}
\end{equation}

\section{DRIVING BEHAVIOUR MODELLING}
\label{sec:modelling}
As discussed above, although EUT has been extensively utilized to model human driving behaviors, there are substantial evidences showing that actual human behaviors often systematically deviate from it. Motivated by this, we reformulate the decision-making process of human drivers in the framework of the CPT model, aiming to capture some irrational behaviors/decisions of human drivers under uncertainties in interactive driving scenarios.
\subsection{Modelling the decision-making process via CPT}
\label{subsec:CPT_modelling}
We consider the driving scenarios with two interacting drivers. Each driver has two discrete decisions/actions: yield and pass. Such scenario can be found in many urban driving circumstances such as intersections, roundabouts and ramp merging. 

Throughout the paper, we refer the predicted vehicle as the target vehicle (denoted with subscript $(\cdot)_T$), and the other one as the interacting vehicle (with $(\cdot)_I$). Denote the action set with pass and yield as $\{\bm{a}\}{=}\{a_{p}, a_{y}\}$. At time $t$, given the historical trajectories of both vehicles, $\{\xi^t_I, \xi^t_T\}$, we aim to obtain an interpretable decision-making model to predict the decision of the target vehicle. 
Note that in interactive driving scenarios, the responses from the interacting vehicle are probabilistic in nature, which will bring uncertainties to the decision-making process of the target vehicles. Under the decision $a_{p}$, the target vehicle has to consider the possibility of the interacting vehicle not yielding, which might force the target vehicle to brake and fail to pass. For the decision $a_{y}$, however, we can assume that it will always succeed. Hence, the prospects for $ a_{p}$ and $a_{y}$ are, respectively,
\begin{IEEEeqnarray}{rCl}
 	P_{a_p}&{=}&\left\{\left(u(\hat{\xi}_{I,y}, \hat{\xi}_{T, p}), p_{I,y}\right){,}\left(u(\hat{\xi}_{I,ny}, \hat{\xi}_{T, p}), 1{-}p_{I,y}\right)\right\},\qquad\\
 	P_{a_y}&{=}&\left\{\left(u(\hat{\xi}_{I,ny}, \hat{\xi}_{T, y}), 1.0\right)\right\}.\label{eq:prospects}
\end{IEEEeqnarray}
where $p_{I,y}$ represents the probability of the interacting vehicle yielding to the target one, and $\hat{\xi}_{I,y}$ and $\hat{\xi}_{I,ny}$ are, respectively, the yielding and non-yielding trajectories of the interacting vehicle. Similarly, $\hat{\xi}_{T,p}$ and $\hat{\xi}_{T,y}$ are, respectively, the passing and yielding trajectories of the target vehicle.

Set $u_0{=}0$. Recalling the CPT model defined in (\ref{eq:CPT})-(\ref{eq:functional_form_w}), we can write the CPT values of the target vehicle under different decisions as:
 \begin{IEEEeqnarray}{rCl}
	V(a_{p}) &{=}& v\left(u^+(\hat{\xi}_{I,y}, \hat{\xi}_{T, p})\right)\left(w^+(1.0){-}w^+(p_{I,y})\right)\nonumber\\
	&&+v\left(u^+(\hat{\xi}_{I,ny}, \hat{\xi}_{T, p})\right)w^+(p_{I,y})\nonumber\\
	&{=}&\left(u(\hat{\xi}_{I,y}, \hat{\xi}_{T, p})\right)^\alpha\left(1{-} \dfrac{p_{I,y}^\gamma}{\left(p_{I,y}^\gamma+(1-p_{I,y})^\gamma\right)^{1/\gamma}}\right)\nonumber\\
	&&+\left(u(\hat{\xi}_{I,ny}, \hat{\xi}_{T, p})\right)^\alpha \dfrac{p_{I,y}^\gamma}{\left(p_{I,y}^\gamma+(1-p_{I,y})^\gamma\right)^{1/\gamma}},\label{eq:decision_making_formulation_1}\\
	V(a_{y}) &{=}& v\left(u^+(\hat{\xi}_{I,ny}, \hat{\xi}_{T, y})\right)w^+(1.0){=}\left(u(\hat{\xi}_{I,ny}, \hat{\xi}_{T, y})\right)^\alpha.\nonumber\\\label{eq:decision_making_formulation_2}
\end{IEEEeqnarray}
Note that in (\ref{eq:decision_making_formulation_1})-(\ref{eq:decision_making_formulation_2}), $u_0{=}0$ simplifies $u(\cdot)$ to $u^+(\cdot)$. 

The decision of the target vehicle is then written as
\begin{equation}
a^* = \arg \max_{a\in\{a_p, a_y\}}\{V(a_p), V(a_y)\}.\label{eq:CPT-decision_driving}
\end{equation}

\subsection{Hierarchical learning of the model parameters}
In the CPT-based decision-making model given in (\ref{eq:decision_making_formulation_1})-(\ref{eq:decision_making_formulation_2}), we have many unknowns that need to be learned from data: the parameters $\alpha$ and $\gamma$, the utility function $u(\cdot)$, and the probability $p_{I,y}$ given $\{\xi_I^t, \xi_T^t\}$. We propose to learn them hierarchically based on the following two assumptions:

\noindent\textbf{Assumption} 1: The parametrization of the utility function of the target vehicle $u{:}(\xi_I, \xi_T){\rightarrow}\mathcal{R}$ does not change with decisions. For instance, if we assume that $u$ is a linear combination of a set of features defined on trajectories, the weights of the features will not change.

\noindent\textbf{Assumption} 2: When the target vehicle is evaluating the CPT value under each decision, the best achievable utilities corresponding to different responses of the interacting vehicle will be adopted. Namely, in (\ref{eq:decision_making_formulation_1})-(\ref{eq:decision_making_formulation_2}), we assume $u(\hat{\xi}_{I,y}, \hat{\xi}_{T, p}){\approx}u(\hat{\xi}_{I,y}, \xi^*_{T, p}(\hat{\xi}_{I,y}))$, $u(\hat{\xi}_{I,ny}, \hat{\xi}_{T, p})\approx u(\hat{\xi}_{I,ny}, \xi^*_{T, p}(\hat{\xi}_{I,ny}))$, and $u(\hat{\xi}_{I,ny}, \hat{\xi}_{T, y})\approx u(\hat{\xi}_{I,ny}, \xi^*_{T, y}(\hat{\xi}_{I,ny}))$.
\vspace{3pt}
\subsubsection{Learning the utility function} We start with learning the utility function $u$ for the target vehicle. With Assumption 1, we learn $u$ from a set of decision-free trajectories of the target vehicle, so that influences of decisions on the demonstrated trajectories can be avoided. This transforms the learning of the utility function into a typical IRL problem. We assume that $u$ is a linear combination of a set of selected features $\bm{\phi}=[\phi_1, \phi_2, {\cdots}, \phi_M]$ defined over $(\xi^{1:N}_I, \xi^{1:N}_T)$ with a horizon length of $N$:
\begin{IEEEeqnarray}{rCl}
	u(\xi^{1:N}_I, \xi^{1:N}_T;\bm{\theta})&=&\bm{\theta}^T \sum_{k=1}^{N}\bm{\phi}(\xi^{k}_I, \xi^{k}_T)\label{eq:linear_cost}.
\end{IEEEeqnarray}
The goal is to find the weights $\bm{\theta}$ which maximizes the likelihood of the demonstration set $\mathcal{U}_{D}{:=}\{(\xi^{1:N}_{I,i}, \xi^{1:N}_{T, i}), i{=}1,{\cdots}, \vert\mathcal{U}_D\vert\}$:
\begin{equation}
\theta^*=\arg\max_{\bm{\theta}}P(\mathcal{U}_{D}|\bm{\theta}).\label{eq:optimal_lambda}
\end{equation}
With the principle of maximum entropy \cite{ziebart2008maximum}, trajectories with higher utilities are exponentially more likely:
\begin{equation}
{P(\xi^{1:N}_{T, i}, \bm{\theta}| \xi^{1:N}_{I,i})} \propto \exp\left(u(\xi^{1:N}_{T, i}, \xi^{1:N}_{I,i};\bm{\theta})\right).
\end{equation}
Thus (\ref{eq:optimal_lambda}) becomes 
\begin{IEEEeqnarray}{rCl}
\bm{\theta}^*&=&\arg\max_{\bm{\theta}}P(\mathcal{U}_{D}|\bm{\theta}){=}\arg\max_{\bm{\theta}}\Pi_{i=1}^{\vert\mathcal{U}_D\vert}\dfrac{P(\xi^{1:N}_{T, i}, \bm{\theta}| \xi^{1:N}_{I,i})}{P(\bm{\theta})}\nonumber\\
&{=}&
\arg\max_{\bm{\theta}}\Pi_{i=1}^{\vert\mathcal{U}_D\vert}\dfrac{P(\xi^{1:N}_{T, i}, \bm{\theta}| \xi^{1:N}_{I,i})}{\int {P(\tilde{\xi}^{1:N}_{T}, \bm{\theta}| \xi^{1:N}_{I,i})}d\tilde{\xi}^{1:N}_{T}}.\label{eq:maximum_entropy}   
\end{IEEEeqnarray}
To solve (\ref{eq:maximum_entropy}), we use the continuous-domain IRL algorithm proposed in \cite{Levine2012ICML}. One can refer \cite{Levine2012ICML} for details.

\subsubsection{Evaluating the utilities and probabilities}
Once the utility function $u$ is obtained, we can generate the best achievable utilities under different decisions. Based on Assumption 2, utilities under different decisions are generated as follows:
\begin{itemize}
	\item $u(\hat{\xi}_{I,y}, \hat{\xi}_{T, p})$ describes the utility when the target vehicle passes, and the interacting vehicle yields. It can be approximated by $u(\hat{\xi}_{I,y}, \xi^*_{T, p}(\hat{\xi}_{I,y}))$ with $\small \xi^*_{T, p}(\hat{\xi}_{I,y}){=}{\arg}{\max}_{\xi_{T,p}}u\left(\hat{\xi}_{I,y}, \xi_{T,p}\right)$. Intuitively, this utility is equivalent to the best achievable utility as if the interacting vehicle was not there since it would yield to the target vehicle.
	\item $u(\hat{\xi}_{I,ny}, \hat{\xi}_{T, p})$, on the other hand, describes the utility when the target vehicle passes but with a non-yielding interacting vehicle. Under this situation, the target vehicle might have to brake and terminate the action of passing. Therefore, we set $\hat{\xi}_{I,ny}$ as if the interacting vehicle was maintaining its initial speed. $\xi^*_{T, p}(\hat{\xi}_{I,ny})$ is calculated as $\xi^*_{T, p}(\hat{\xi}_{I,ny})=[\xi^{*, 1:k_0}_{T, p}(\hat{\xi}_{I,y}); \xi^{k_0:N}_{T, brake}]$ where the first part $\xi^{*, 1:k_0}_{T, p}(\hat{\xi}_{I,y})$ is the first $k_0$ steps of an optimal passing trajectory, while the second part $\xi^{k_0:N}_{T, brake}$ is a braking trajectory in order to avoid collision with the interacting vehicle. The maximum value of $k_0$ is found via boundaries on deceleration. Hence, the corresponding utility in this situation is given by $u\left(\hat{\xi}_{I,ny}, [\xi^{*, 1:k_0}_{T, p}(\hat{\xi}_{I,y}); \xi^{k_0:N}_{T, brake}]\right)$.
	\item $u(\hat{\xi}_{I,ny}, \hat{\xi}_{T, y})$ is the utility if the target vehicle chooses to yield. For this scenario, it does not matter whether the interacting vehicle yields or not. We can directly solve for the optimal trajectory for the target vehicle with additional constraints on its trajectories. For instance, we can set an upper bound for the achievable zones of its trajectories to force it to yield \footnote[1]{The settings of the upper bound differ depending on scenarios. For interactions and roundabouts, the upper bound comes from the traffic-rule maps such as the locations of stop bars. For ramp merging, the upper bound can be draw from the trajectory of the interacting vehicle. We will explain more details about this in the case study.}. Hence, $u(\hat{\xi}_{I,ny}, \hat{\xi}_{T, y})= \min_{{\xi}_{T, y}}u(\hat{\xi}_{I,ny}, \xi_{T, y}(\hat{\xi}_{I,ny}))$ with constraints in the form of $g({\xi}_{T, y})\le0$.
\end{itemize}

Apart from the utilities, we also need to find an objective probability variable that can quantify approximately how the interacting vehicle will respond if the target vehicle was to take the pass action, i.e., approximating $p_{I,y}$ in (\ref{eq:decision_making_formulation_1}) given historical observations $(\xi^t_{I}, \xi^t_{T})$. Inspired by the predefined models, we use the TTC to approximate it. Define the TTC gap between the two interacting vehicles as $\triangle_{TTC}{=}TTC_{T}{-}TTC_{I}$. We assume that $p_{I,y}$ is higher if $\triangle_{TTC}$ is lower:
\begin{equation}
	p_{I,y} = \dfrac{1}{1+\exp\left(TTC_{T}-TTC_{I} \right)}\label{eq:probability_approximation}
\end{equation} 

\subsubsection{Learning the value function and decision weighting function}
With the acquired utilities and probabilities, the next step is to formulate a learning problem to find the unknown parameter $\alpha$ in the value function as well as $\gamma$ in the decision weighting function in (\ref{eq:decision_making_formulation_1})-(\ref{eq:decision_making_formulation_2}). To achieve this goal, we again adopt the principle of maximum entropy to convert the decision selection process in (\ref{eq:CPT-decision_driving}) as a soft-max problem:
\begin{IEEEeqnarray}{rCl}
	Pr(a_p) = \dfrac{1}{1+\exp\left(V(a_y){-}V(a_p)\right)}\label{eq:soft-max-1},\\
	Pr(a_y) = \dfrac{1}{1+\exp\left(V(a_p){-}V(a_y)\right)}\label{eq:soft-max-2}.
\end{IEEEeqnarray}
where $Pr(a_p)$ and $Pr(a_y)$ represent the probabilities of choosing action $a_p$ and $a_y$, respectively. 
Given a set of $K$ interactive trajectories with labelled decisions for the target vehicle, denoted by $S = \{(\xi^i_I, \xi^i_T, a^i_T), i{=}1,{\cdots}, K\}$, we can formulate the learning of $\alpha$ and $\gamma$ as a nonlinear logistic regression problem with the loss function as:
\begin{IEEEeqnarray}{rCl}
	L(\alpha, \gamma)&{=}&\sum_{i}^K\left\{{-}\bm{1}(a^i_T{=}a_p)\log Pr_i(a_p)\right.\label{eq:regression}\\
	&&\qquad\left.{-}\bm{1}(a^i_T{=}a_y)\log\left(1{-}Pr_i(a_y)\right) \right\}\nonumber,
\end{IEEEeqnarray}
where $\bm{1}(x){=}1$ if $x=1$ and $\bm{1}(x){=}0$ otherwise. $Pr_i(a_y)$ and $Pr_i(a_p)$ are the evaluated probabilities as in (\ref{eq:soft-max-1})-(\ref{eq:soft-max-2}) on the $i$-th pair of interactive trajectories $(\xi^i_I, \xi^i_T, a^i_T)$ based on (\ref{eq:decision_making_formulation_1})-(\ref{eq:decision_making_formulation_2}) and (\ref{eq:probability_approximation}). 

The optimal $\alpha$ and $\gamma$ can be found via
\begin{equation}
	(\alpha^*, \gamma^*) = \arg\min_{\alpha, \gamma}L(\alpha, \gamma).\label{eq:optimal_al_gamma}
\end{equation}

With the three steps described above, all the unknowns in the CPT model in (\ref{eq:decision_making_formulation_1})-(\ref{eq:decision_making_formulation_2}) can be obtained. 

\section{CASE STUDY}
\label{sec:case_study}
\subsection{A driving scenario: roundabout}
\begin{figure}[h!]
	\begin{center}
		\includegraphics[width=\linewidth]{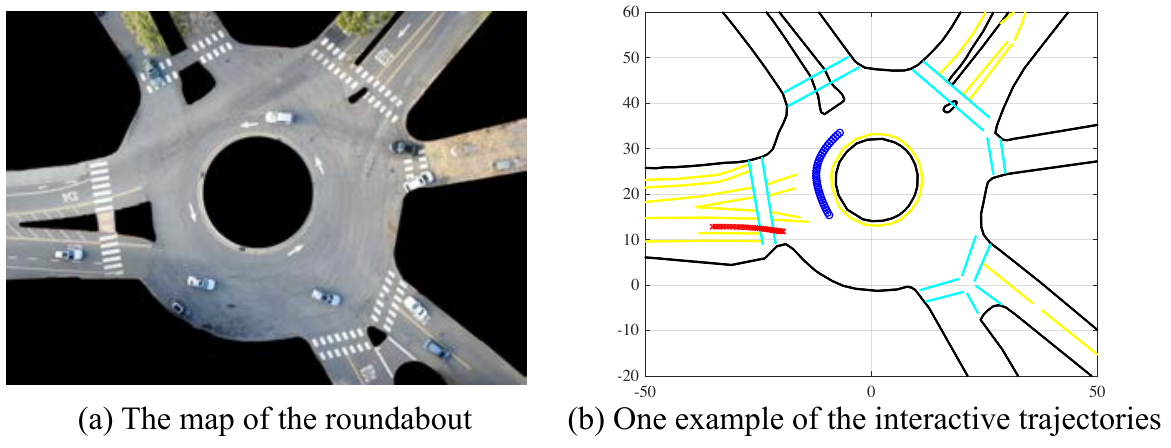}
		\caption{The map of the roundabout and one example pair of interactive trajectories. Red stars: the target vehicle; Blue circles: the interacting vehicle.}
		\label{fig: roundabout}
	\end{center}
\end{figure}

To evaluate the performance of the proposed approach, we select a roundabout merging scenario from the INTERACTION dataset \cite{Wei2019IROS, zhan_2019}. As shown in \Cref{fig: roundabout}(a), the roundabout has 6 branches and each branch has two directions (both in and out). We selected the interactive motions of two cars at the left-most branch (\Cref{fig: roundabout}(b)) because there is no enforced stop signs at this branch before merging into the roundabout. This makes the interaction more intensive, and consequently creating more challenging problems.

We define the merging-in vehicle (the blue one in \Cref{fig: roundabout}(b)) as the target vehicle, and the one already in the roundabout as the interacting vehicle (the red one in \Cref{fig: roundabout}(b)). Based on a period of historical data on both vehicles, different driving behavior models try to predict whether the red target vehicle will decide to merge in front of the interacting vehicle in blue (i.e, the target vehicle passes), or wait to merge in until the blue car passes (i.e., the target vehicle yields).

\begin{figure}[h!]
	\begin{center}
		\includegraphics[width=\linewidth]{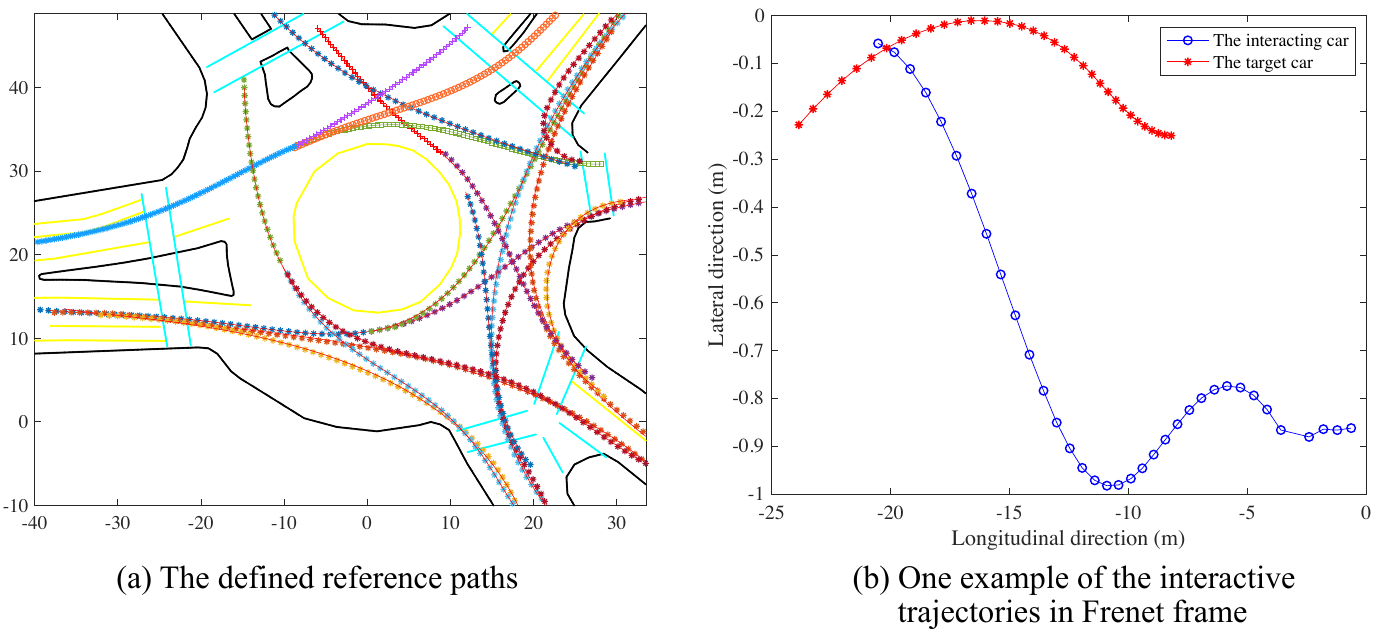}
		\caption{The reference paths (a) and trajectories in Frenet frame (b). The crossing points on the pair of reference paths define the common reference zero points for two interactive cars.}
		\label{fig: references_and_example_frenet}
	\end{center}
\end{figure}

We use the Frenet frame \cite{wang_modeling_2014} to represent the trajectory coordinates of each vehicle. Reference paths of the map are shown in \Cref{fig: references_and_example_frenet}(a). To capture the relationships between the two cars on the longitudinal direction, we set the crossing point of the reference paths of the two interactive cars as their shared reference point. Before the crossing point, the longitudinal coordinates of both cars are negative. Once passed the crossing point, both longitudinal coordinates become positive. One example of the interactive trajectories in the defined Frenet frame can be found in \Cref{fig: references_and_example_frenet}(b).
\subsection{Comparison models}  
We compared the decision prediction performance among three different models: 1) a predefined TTC rule-based model, 2) a learning-based neural network model, and 3) the proposed CPT model. A brief introduction of each model is given below.
\vspace{1mm}
\subsubsection{The TTC rule-based model}
The TTC rule-based model uses the TTC as an indicator to predict which car will go first between the two interactive cars. Given the trajectories of each car in Frenet frame as shown in \Cref{fig: references_and_example_frenet}(b), the TTC can be easily calculated via 
\begin{equation}
	TTC^t_{I,T} = s^t_{I,T}/v^t_{I,T}\label{eq:TTC_cal}
\end{equation}
where $s^t_{I,T}$ represents the longitudinal length from the current location of the cars at time $t$ to the collision point along the reference paths. $v^t_{I,T}$ is the current speed of the cars.

As discussed in (\ref{eq:probability_approximation}) in \Cref{sec:modelling}, we calculate the soft-max probability of the target car passing via
\begin{equation}
Pr^t(a_p) = \dfrac{1}{1+\exp\left(TTC^t_{T}-TTC^t_{I} \right)}.\label{eq:TTC_model}
\end{equation} 
\subsubsection{The learning-based neural network model}
The learning-based model we used is based on neural networks (NNs). The input is a period of historical trajectories of the two interacting vehicles in Frenet frame. The first layer is a long short-term memory (LSTM) cell with 16 neurons, followed by two fully connected layers, and each with 8 neurons. Afterwards, a \textit{tanh} nonlinear activation layer is applied, with a the softmax layer as the final layer to output the classification results. In order to avoid over-fitting, we applied the drop-out technique to the fully connected layers with a dropout rate of 0.5 and added a L2 regularization term to the original cross-entropy loss function.
\vspace{2mm}
\subsubsection{The proposed CPT model}
In the proposed CPT model, we have selected four features in the utility function. They are defined as:
\begin{itemize}
	\item Speed feature $\phi_1{=}\exp\left(-(v^t-v_{traffic})^2\right)$;
	\item Acceleration feature $\phi_2{=}\exp\left(-(acc^t)^2\right)$;
	\item Jerk feature $\phi_3{=}\exp\left(-(jerk^t)^2\right)$;
	\item Safety feature $\phi_4{=}\exp\left((s^t_I-s^t_T)^2\right)$.
\end{itemize}
Note that all the variables $v^t$, $acc^t$ and $jerk^t$ can be written as linear functions of the trajectories of the target vehicle based on backward differentiation.

\begin{figure}[h!]
	\begin{center}
		\includegraphics[width=\linewidth]{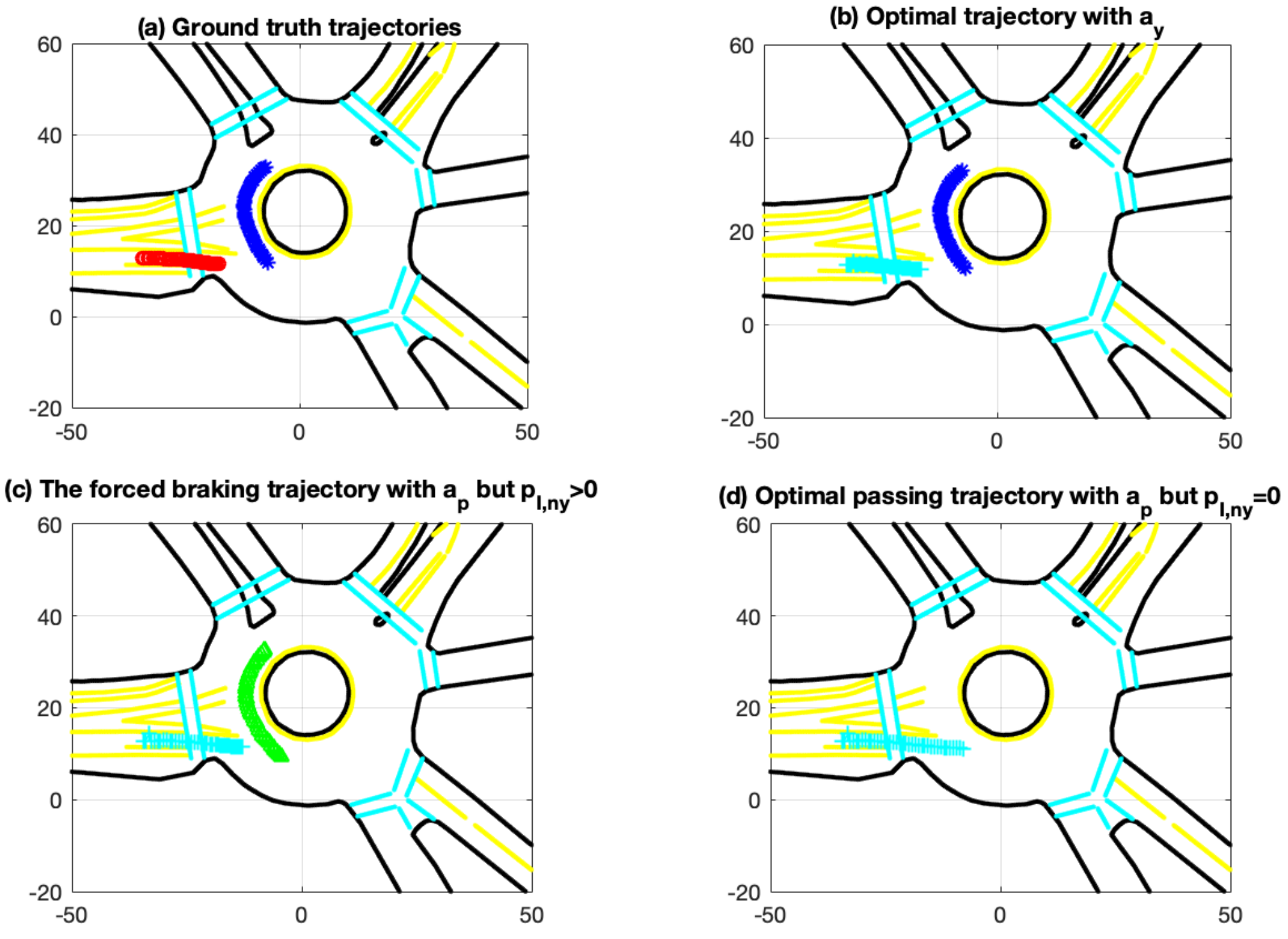}
		\caption{An example of the trajectories used for utility calculation under different decisions and different responses of the interacting vehicle: (a) the ground truth trajectories (red: the target car; blue: the interacting car); (b) the optimal trajectory of the target car (cyan) under the decision of yielding ($a_p$), and the ground truth of the interacting car (blue); (c) the forced braking trajectory of the target car (cyan) under a passing decision but with a non-yielding interacting car (green). The virtual trajectory of the interacting car is assumed to maintain its initial speed; (d) the optimal trajectory of the target car (cyan) under a passing decision with a yielding interacting car.}
		\label{fig: utility_trajs}
	\end{center}
\end{figure}

As for the calculation of key utilities in (\ref{eq:decision_making_formulation_1})-(\ref{eq:decision_making_formulation_2}), examples of the corresponding trajectories for the utility evaluation are shown in \Cref{fig: utility_trajs}. The ground truth interactive trajectories are shown in \Cref{fig: utility_trajs}(a) with red for the target vehicle and blue for the interacting vehicle. Figure \ref{fig: utility_trajs}(b) shows the optimal yielding trajectory of the target vehicle (cyan) and the ground truth trajectory of the interacting vehicle (blue). Figure \ref{fig: utility_trajs}(c) and \Cref{fig: utility_trajs}(d), respectively, show the trajectories of the target vehicle (cyan) under a passing decision with a non-yielding and yielding interacting vehicles. If the interacting is not-yielding, we assume that it will maintain its initial speed, as shown in green in \Cref{fig: utility_trajs}(c). In this case, the target vehicle is forced to brake. On the other hand, with a yielding interacting vehicle, the optimal passing trajectory of the target vehicle is shown in \Cref{fig: utility_trajs}(d).

\subsection{Experiment results and discussion}
We discuss the experimental results in two aspects: prediction performance comparison among the three models, and the interpretability of these models.
\vspace{1mm}
\subsubsection{Comparison of the prediction performance}
We trained and tested all three models on a dataset containing 67 pairs of interacting trajectories with a sampling frequency of 10Hz. To learn more generalized results, we slice the trajectories into frames with a fixed length using moving windows. Each frame contains the trajectories in 1s. Thus, all 67 pairs of interacting trajectories generate 2680 frames. To achieve better performance for the learning-based model, we have conducted two sets of experiments for the training of the neural network: 
\begin{itemize}
	\item Experiment 1: randomly shuffle all the trajectory pairs and select 80\% of them for training and the other 20\% for testing. The success rate \footnote[2]{Success rate is defined as the percentage of correct predictions among all test examples.} is 65\% for testing.
	\item Experiment 2: directly shuffle all frames for the neural network and randomly select 80\% for training and 20\% for testing. The success rate is 97\% for testing.
\end{itemize}
The large discrepancy between the testing accuracies of the two experiments with the NN model is mainly due to the over-fitting problem cause by the data insufficiency. In experiment 1, it showed that the NN model learned on 80\% of the trajectory pairs cannot be well generalized to other interaction pairs. 

We list the success rates for prediction from all three models in \Cref{table: 1}. It shows that the proposed CPT model outperformed the TTC model and the NN model in experiment 1, and it achieved similar performance as the NN model in experiment 2. Moreover, both the TTC model and the proposed CPT model are more data-efficient for similar achievable performance.
\vspace{2mm}
\begin{table}[h!]
	\caption{Comparison of the success rates in three models}\centering{}%
	\begin{tabular}{c|cccc}
		& \makecell{TTC} & \makecell{NN} & \makecell{CPT} \\ \hline
		\makecell{Success rates} & 81.82\% & \makecell{97\%} & 95.45\%  \\ \hline
	\end{tabular}%
	\label{table: 1}
\end{table}

\subsubsection{Interpretability of the CPT model}
\begin{figure}[h!]
	\begin{center}
		\includegraphics[width=0.8\linewidth]{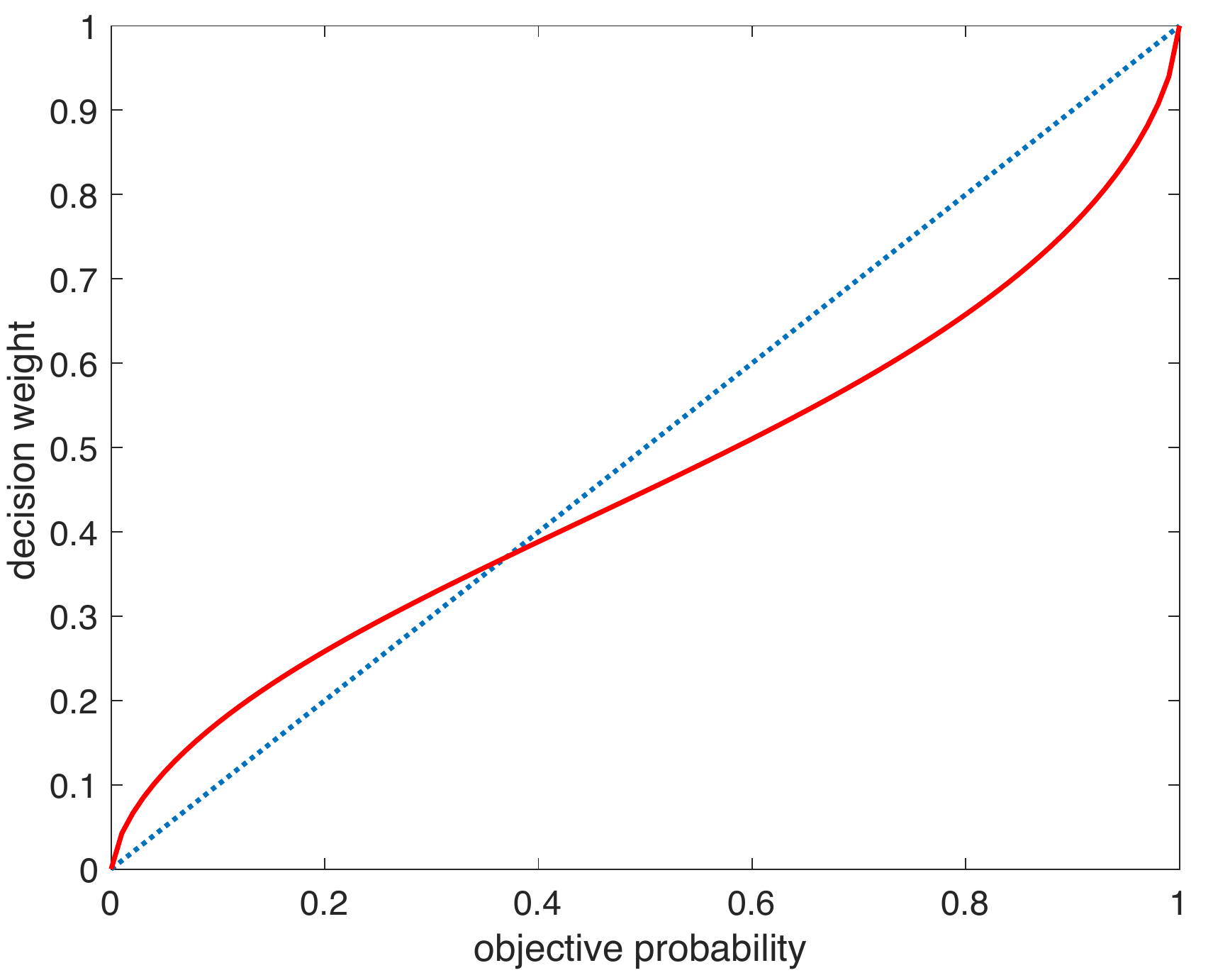}
		\caption{The learned decision weighting function (red curve)}
		\label{fig: decision_weights}
	\end{center}
\end{figure}

In the CPT model, the parameters we have learned via the nonlinear logistic regression are 
\begin{equation}
\alpha^* = 0.9827, \gamma^* = 0.6742. \label{eq: value_al_g}
\end{equation}
With the optimal $\gamma$, the learned decision weighting function is shown in \Cref{fig: decision_weights}. We can see that the CPT model indeed captured the human choice patterns that events with low probabilities will tend to be overestimated, while high-probability events are often underestimated. Such results are consistent with many studies about human behavior in other domains such as economics, investment and waiting paradox problems.

\section{CONCLUSION}
In this paper, we proposed an interpretable and irrationality-aware human behavior model in interactive driving scenarios based on the cumulative prospect theory (CPT). To learn the model parameters from real driving data, a hierarchical learning algorithm was also developed, in which inverse reinforcement learning and nonlinear logistic regression were combined. Comparison studies were conducted among three different models: a predefined TTC model, a neural network (NN) based learning model, and the proposed CPT model. The results showed that the proposed CPT model outperformed the TTC model in terms of prediction accuracy. Similar performance was achieved by the CPT model as the NN model, but with much less amount of data. Moreover, the learned parameters of the CPT model have explicit and interpretable physical meanings, which matched the observations of the human behavior in many domains.

\bibliographystyle{IEEEtran}
\bibliography{ITSC2019}

\begin{thebibliography}{10}
\providecommand{\url}[1]{#1}
\csname url@rmstyle\endcsname
\providecommand{\newblock}{\relax}
\providecommand{\bibinfo}[2]{#2}
\providecommand\BIBentrySTDinterwordspacing{\spaceskip=0pt\relax}
\providecommand\BIBentryALTinterwordstretchfactor{4}
\providecommand\BIBentryALTinterwordspacing{\spaceskip=\fontdimen2\font plus
\BIBentryALTinterwordstretchfactor\fontdimen3\font minus
  \fontdimen4\font\relax}
\providecommand\BIBforeignlanguage[2]{{%
\expandafter\ifx\csname l@#1\endcsname\relax
\typeout{** WARNING: IEEEtran.bst: No hyphenation pattern has been}%
\typeout{** loaded for the language `#1'. Using the pattern for}%
\typeout{** the default language instead.}%
\else
\language=\csname l@#1\endcsname
\fi
#2}}

\bibitem{abuali_driver_2016}
N.~AbuAli and H.~Abou-zeid, ``Driver {Behavior} {Modeling}: {Developments} and
  {Future} {Directions},'' \emph{International Journal of Vehicular
  Technology}, 2016.

\bibitem{wang_modeling_2014}
W.~Wang, J.~Xi, and H.~Chen, ``Modeling and {Recognizing} {Driver} {Behavior}
  {Based} on {Driving} {Data}: {A} {Survey},'' \emph{Mathematical Problems in
  Engineering}, 2014.

\bibitem{rahman_review_2013}
M.~Rahman, M.~Chowdhury, Y.~Xie, and Y.~He, ``Review of {Microscopic}
  {Lane}-{Changing} {Models} and {Future} {Research} {Opportunities},''
  \emph{IEEE Transactions on Intelligent Transportation Systems}, vol.~14,
  no.~4, pp. 1942--1956, Dec. 2013.

\bibitem{mcdonald_development_1997}
M.~McDonald, J.~Wu, and M.~Brackstone, ``{Development of a Fuzzy Logic based
  Microscopic Motorway Simulation Model},'' in \emph{Proceedings of
  {Conference} on {Intelligent} {Transportation} {Systems}}.\hskip 1em plus
  0.5em minus 0.4em\relax Boston, MA, USA: IEEE, 1997, pp. 82--87.

\bibitem{kiefer_status_2005}
R.~J. Kiefer, J.~Salinger, and J.~J. Ference, ``Status of {NHTSA}'s
  {Rear}-{End} {Crash} {Prevention} {Research} {Program},'' June 2005.

\bibitem{treiber_congested_2000}
M.~Treiber, A.~Hennecke, and D.~Helbing, ``Congested {Traffic} {States} in
  {Empirical} {Observations} and {Microscopic} {Simulations},'' \emph{Physical
  Review E}, vol.~62, no.~2, pp. 1805--1824, Aug. 2000.

\bibitem{kesting_general_2007}
A.~Kesting, M.~Treiber, and D.~Helbing, ``General {Lane}-{Changing} {Model}
  {MOBIL} for {Car}-{Following} {Models},'' \emph{Transportation Research
  Record}, vol. 1999, no.~1, pp. 86--94, Jan. 2007.

\bibitem{aoude_driver_2012}
G.~S. Aoude, V.~R. Desaraju, L.~H. Stephens, and J.~P. How, ``Driver {Behavior}
  {Classification} at {Intersections} and {Validation} on {Large}
  {Naturalistic} {Data} {Set},'' \emph{IEEE Transactions on Intelligent
  Transportation Systems}, vol.~13, no.~2, pp. 724--736, June 2012.

\bibitem{hu_probabilistic_2018}
Y.~Hu, W.~Zhan, and M.~Tomizuka, ``Probabilistic {Prediction} of {Vehicle}
  {Semantic} {Intention} and {Motion},'' in \emph{2018 {IEEE} {Intelligent}
  {Vehicles} {Symposium} ({IV})}, June 2018, pp. 307--313.

\bibitem{li_generic_2018IV}
J.~Li, W.~Zhan, and M.~Tomizuka, ``Generic {Vehicle} {Tracking} {Framework}
  {Capable} of {Handling} {Occlusions} {Based} on {Modified} {Mixture}
  {Particle} {Filter},'' in \emph{2018 {IEEE} {Intelligent} {Vehicles}
  {Symposium} ({IV})}, June 2018, pp. 936--942.

\bibitem{li_coordination_2019}
J.~Li, H.~Ma, W.~Zhan, and M.~Tomizuka, ``Coordination and {Trajectory}
  {Prediction} for {Vehicle} {Interactions} via {Bayesian} {Generative}
  {Modeling},'' in \emph{{IEEE} {Intelligent} {Vehicles} {Symposium}}, 2019.

\bibitem{gupta2018social}
A.~Gupta, J.~Johnson, L.~Fei-Fei, S.~Savarese, and A.~Alahi, ``Social gan:
  Socially acceptable trajectories with generative adversarial networks,'' in
  \emph{Proceedings of the IEEE Conference on Computer Vision and Pattern
  Recognition}, 2018, pp. 2255--2264.

\bibitem{hu_framework_2018}
Y.~Hu, W.~Zhan, and M.~Tomizuka, ``A {Framework} for {Probabilistic} {Generic}
  {Traffic} {Scene} {Prediction},'' in \emph{2018 21st {International}
  {Conference} on {Intelligent} {Transportation} {Systems} ({ITSC})}, Nov.
  2018, pp. 2790--2796.

\bibitem{ma2018trafficpredict}
Y.~Ma, X.~Zhu, S.~Zhang, R.~Yang, W.~Wang, and D.~Manocha, ``Trafficpredict:
  Trajectory prediction for heterogeneous traffic-agents,'' \emph{arXiv
  preprint arXiv:1811.02146}, 2018.

\bibitem{Hu2019IV}
Y.~Hu, W.~Zhan, L.~Sun, and M.~Tomizuka, ``Multi-modal probabilistic prediction
  of interactive behavior via an interpretable model,'' in \emph{Proceedings of
  the IEEE Intelligent Vehicle Symposium (IV2019)}, 2019.

\bibitem{premack1978does}
D.~Premack and G.~Woodruff, ``Does the chimpanzee have a theory of mind?''
  \emph{Behavioral and brain sciences}, vol.~1, no.~4, pp. 515--526, 1978.

\bibitem{baker2014modeling}
C.~L. Baker and J.~B. Tenenbaum, ``Modeling human plan recognition using
  bayesian theory of mind,'' \emph{Plan, activity, and intent recognition:
  Theory and practice}, pp. 177--204, 2014.

\bibitem{driggs-campbell_integrating_2017}
K.~Driggs-Campbell, V.~Govindarajan, and R.~Bajcsy, ``Integrating {Intuitive}
  {Driver} {Models} in {Autonomous} {Planning} for {Interactive} {Maneuvers},''
  \emph{IEEE Transactions on Intelligent Transportation Systems}, vol.~18,
  no.~12, pp. 3461--3472, Dec. 2017.

\bibitem{abbeel2004apprenticeship}
P.~Abbeel and A.~Y. Ng, ``Apprenticeship learning via inverse reinforcement
  learning,'' in \emph{Proceedings of the twenty-first international conference
  on Machine learning}.\hskip 1em plus 0.5em minus 0.4em\relax ACM, 2004, p.~1.

\bibitem{ziebart2008maximum}
B.~D. Ziebart, A.~L. Maas, J.~A. Bagnell, and A.~K. Dey, ``Maximum entropy
  inverse reinforcement learning.'' in \emph{AAAI}, vol.~8.\hskip 1em plus
  0.5em minus 0.4em\relax Chicago, IL, USA, 2008, pp. 1433--1438.

\bibitem{Levine2012ICML}
S.~Levine and V.~Koltun, ``continuous inverse optimal control with locally
  optimal examples,,'' in \emph{the 29th International Conference on Machine
  Learning (ICML-12)}, 2012.

\bibitem{sun_courteous_2018}
L.~Sun, W.~Zhan, M.~Tomizuka, and A.~D. Dragan, ``Courteous {Autonomous}
  {Cars},'' in \emph{2018 {IEEE}/{RSJ} {International} {Conference} on
  {Intelligent} {Robots} and {Systems} ({IROS})}, Oct. 2018, pp. 663--670.

\bibitem{sun_probabilistic_2018}
L.~Sun, W.~Zhan, and M.~Tomizuka, ``Probabilistic {Prediction} of {Interactive}
  {Driving} {Behavior} via {Hierarchical} {Inverse} {Reinforcement}
  {Learning},'' in \emph{2018 21st {International} {Conference} on
  {Intelligent} {Transportation} {Systems} ({ITSC})}, Nov. 2018, pp.
  2111--2117.

\bibitem{Hu2019ITSC}
Y.~Hu, L.~Sun, and M.~Tomizuka, ``Generic prediction architecture considering
  both rational and irrational driving behavior,'' in \emph{Proceedings of the
  IEEE Transportation System Conference (ITSC2019)}, 2019.

\bibitem{kahneman1979prospect}
D.~Kahneman, ``Prospect theory: An analysis of decisions under risk,''
  \emph{Econometrica}, vol.~47, p. 278, 1979.

\bibitem{tversky1992advances}
A.~Tversky and D.~Kahneman, ``Advances in prospect theory: Cumulative
  representation of uncertainty,'' \emph{Journal of Risk and uncertainty},
  vol.~5, no.~4, pp. 297--323, 1992.

\bibitem{Wei2019IROS}
W.~Zhan, L.~Sun, D.~Wang, Y.~Jin, and M.~Tomizuka, ``{Constructing a Highly
  Interactive Vehicle Motion Dataset},'' in \emph{2019 IEEE/RSJ International
  Conference on Intelligent Robots and Systems (IROS)}, 2019.

\bibitem{zhan_2019}
W.~Zhan, L.~Sun, D.~Wang, H.~Shi, A.~Clausse, M.~Naumann, J.~K\"ummerle,
  H.~K\"onigshof, C.~Stiller, A.~de~La~Fortelle, and M.~Tomizuka,
  ``{INTERACTION Dataset: An INTERnational, Adversarial and Cooperative moTION
  Dataset in Interactive Scenarios with Semantic Maps},'' 2019.

\bibitem{risk1954exposition}
O.~RISK and D.~BERNOULLI, ``Exposition of a new theory on the measurement,''
  \emph{Econometrica}, vol.~22, no.~1, pp. 23--36, 1954.

\bibitem{allais1953rational}
M.~Allais, ``Rational man's behavior in the presence of risk: Critique of the
  postulates and axioms of the american school,'' \emph{Econometrica}, vol.~21,
  no.~4, pp. 503--46, 1953.

\end{thebibliography}
\end{document}